\documentclass[]{bytedance}
\usepackage[toc,page,header]{appendix}

\usepackage{minitoc}
\usepackage{amsfonts}
\usepackage{amssymb}
\usepackage{tabularx}
\usepackage{listings}
\usepackage{xcolor}
\usepackage{cancel}

\usepackage{tabulary,multirow,xspace}
\usepackage{fixmath,mathtools,nicefrac,mmstyle}
\usepackage{subcaption}
\usepackage{caption}
\usepackage{wrapfig} 
\usepackage[misc]{ifsym} 
\usepackage{colortbl}

\usepackage{wrapfig}
\usepackage{multicol}
\usepackage[most]{tcolorbox}
\usepackage{pifont}

\definecolor{codegreen}{rgb}{0,0.6,0}
\definecolor{codegray}{rgb}{0.5,0.5,0.5}
\definecolor{codepurple}{rgb}{0.58,0,0.82}
\definecolor{backcolour}{rgb}{0.95,0.95,0.92}
\definecolor{boxblue}{RGB}{57,89,163}
\definecolor{boxbluebg}{RGB}{230,237,250} 

\lstdefinestyle{mystyle}{
    backgroundcolor=\color{backcolour},   
    commentstyle=\color{codegreen},
    keywordstyle=\color{magenta},
    numberstyle=\tiny\color{codegray},
    stringstyle=\color{codepurple},
    basicstyle=\ttfamily\footnotesize,
    breakatwhitespace=false,         
    breaklines=true,                 
    captionpos=b,                    
    keepspaces=true,                 
    numbers=none,                    
    numbersep=5pt,                  
    showspaces=false,                
    showstringspaces=false,
    showtabs=false,                  
    tabsize=2
}
\definecolor{mygray1}{gray}{.95}
\definecolor{mygray2}{gray}{.9}
\definecolor{mygray3}{gray}{.95}
\usepackage{pifont}

\newcommand{\myparagraph}[1]{{\noindent\bf #1}}

\newlength\savewidth
\newcolumntype{x}[1]{>{\centering\arraybackslash}p{#1pt}}

\newcommand{\app}{\raise.17ex\hbox{$\scriptstyle\sim$}}

\usepackage{xcolor}
\usepackage{graphicx}
\usepackage{amssymb}
\usepackage{pifont}
\usepackage{floatrow}
\usepackage{amsmath} 
\usepackage{float}
\usepackage{wrapfig}
\usepackage{multirow}
\usepackage{tcolorbox}
\tcbuselibrary{breakable, skins, raster}
\usepackage{listings}
\newcommand{\cmark}{{\color{green}\checkmark}}
\newcommand{\xmark}{{\color{red}\ding{55}}}

\usepackage{listings}

\definecolor{commentgreen}{rgb}{0.1, 0.4, 0.1}
\definecolor{keywordblue}{rgb}{0.1, 0.1, 0.7}
\definecolor{stringred}{rgb}{0.7, 0.1, 0.1}

\lstdefinestyle{mystyle}{
    commentstyle=\color{commentgreen},
    keywordstyle=\color{keywordblue},   
    stringstyle=\color{stringred},
    basicstyle=\ttfamily\scriptsize, 
    breaklines=true,
    keepspaces=true,
    showstringspaces=false,
    frame=none,                     
    language=Python, 
}

\newcommand{\name}{Phantom-Data}
\title{\name{}: Towards a General Subject-Consistent Video Generation Dataset}

\author{
\centerline{
Zhuowei Chen $^*$\quad 
    Bingchuan Li $^{*\dagger}$ \quad  
    Tianxiang Ma $^*$ \quad 
    Lijie Liu $^*$ \quad
    Mingcong Liu \quad
} 
\centerline{
    Yi Zhang \quad
    Gen Li  \quad 
    Xinghui Li  \quad 
    Siyu Zhou \quad 
    Qian He  \quad 
    Xinglong Wu  \quad
}
}

\affiliation[]{Intelligent Creation Lab, ByteDance}
\contribution[*]{Equal contribution}
\contribution[\dagger]{Project lead}

\abstract{
Subject-to-video generation has witnessed substantial progress in recent years. However, existing models still face significant challenges in faithfully following textual instructions. 
This limitation, commonly known as the copy-paste problem, arises from the widely used in-pair training paradigm. This approach inherently entangles subject identity with background and contextual attributes by sampling reference images from the same scene as the target video.
To address this issue, we introduce \textbf{Phantom-Data, the first general-purpose cross-pair subject-to-video consistency dataset}, containing approximately one million identity-consistent pairs across diverse categories. Our dataset is constructed via a three-stage pipeline: (1) a general and input-aligned subject detection module, (2) large-scale cross-context subject retrieval from more than 53 million videos and 3 billion images, and (3) prior-guided identity verification to ensure visual consistency under contextual variation. Comprehensive experiments show that training with Phantom-Data significantly improves prompt alignment and visual quality while preserving identity consistency on par with in-pair baselines.
}

\date{\today}

\checkdata[Project Page]{\url{https://phantom-video.github.io/Phantom-Data/}}

\correspondence{Zhuowei Chen, Bingchuan Li at \email{\{chenzhuowei.ustc, libingchuan\}@bytedance.com}}

\begin{document}
\maketitle

\begin{figure}[ht]
  \centering
  \includegraphics[width=\linewidth]{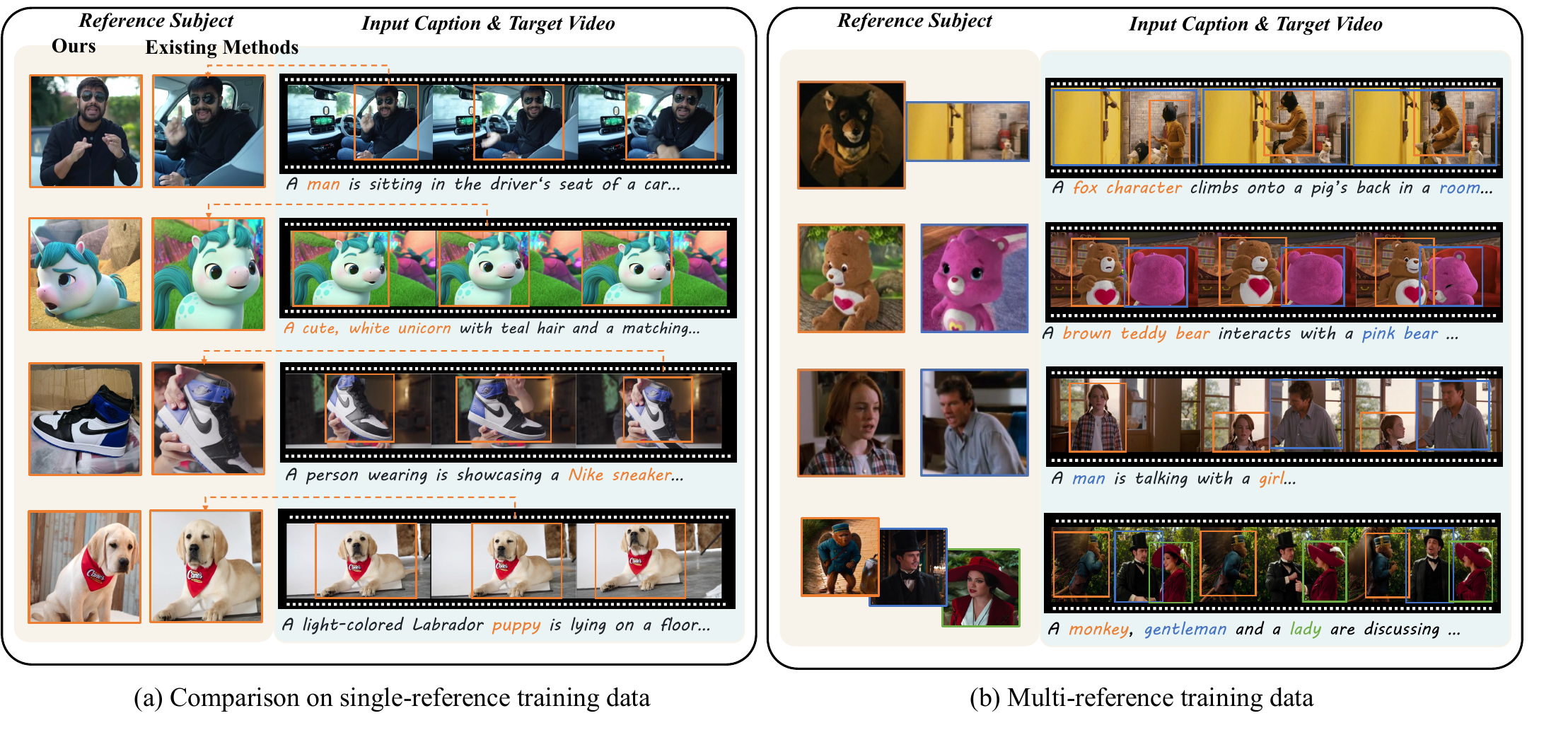}
  \caption{Overview of training samples. (a) Single‑reference setting: existing methods typically extract the reference image from the target video itself. In contrast, our approach uses reference images captured in distinct contexts. (b) Our dataset also includes multi‑reference samples, presenting each subject in varied contextual settings.}
  \label{fig:teaser}
\end{figure}

\section{Introduction}\label{sec:intro}
\vspace{-0.5em}
In recent years, text-to-video generation models, exemplified by Sora \cite{sora}, have made significant progress \cite{seawead2025seaweed, kong2024hunyuanvideo, wan2025, yang2024cogvideox, magi1}. However, due to the limited controllability inherent in textual instructions, achieving fine-grained control over video generation remains a key challenge for practical applications. Among recent advances \cite{hu2024animate, jiang2025vace, zeng2024make, i2vcontrolcamera, kondratyuk2024videopoet, xu2024magicanimate}, increasing attention has been paid to enforcing subject identity consistency in text-to-video generation. The subject-consistent video generation task (S2V) \cite{liu2025phantom, fei2025skyreels, chen2025multi, huang2025conceptmaster, deng2025cinema} aims to generate videos that not only follow the given text prompt but also faithfully preserve the identity of reference subjects, such as people, animals, products, or scenes. This capability has great potential in applications such as personalized advertising \cite{chen2025goku} and AI-driven filmmaking \cite{wu2025automated}.

\begin{figure}[!t]
  \centering
  \includegraphics[width=\linewidth]{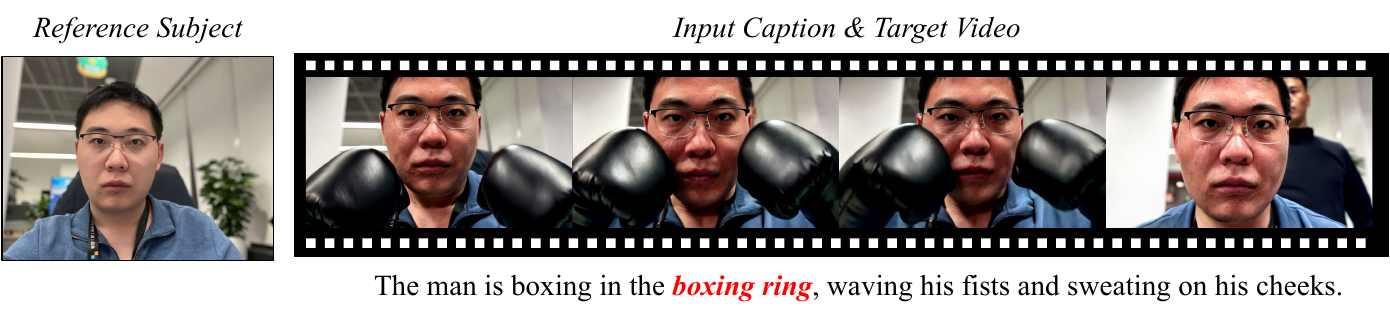}
  \caption{Illustration of the copy-paste problem. The shown result is generated by a SOTA video generation model (Kling \cite{Kling}). 
  }
  \label{fig:copypaste}
\end{figure}

Despite encouraging progress in visual consistency, existing S2V approaches still suffer from limited text-following ability and suboptimal video quality, a phenomenon often referred to as the \textit{copy-paste} problem. As shown in Fig.~\ref{fig:copypaste}, the generated video directly replicates the reference subject from one of its frames, leading to the omission of the "boxing ring" background described in the prompt. This issue stems from the \textit{in-pair} training paradigm, where the reference subject is sampled from the same target video, as illustrated in Fig.\ref{fig:teaser}(a). Consequently \cite{ho2020denoising, liuflow}, the model tends to preserve not only subject identity but also irrelevant contextual details. However, in real-world scenarios, such entangled features may contradict the actions or semantics described in the text prompt, causing the generated videos that either deviate from the prompt or exhibit noticeable artifacts.

To address the above issue, prior works \cite{chen2025multi, liang2025movie, huang2025conceptmaster, jiang2025vace, ju2025fulldit} have explored various data normalization and augmentation strategies, such as background removal, color jittering, and geometric transformations. However, these methods struggle to unravel complex contextual factors, such as viewpoint and motion, due to limited variation.
More recent approaches introduce \textit{cross-pair} data, where identity-consistent reference and target frames are sampled from different sources. This setting encourages the model to focus on identity preservation while reducing overfitting to irrelevant visual contexts \cite{polyak2024movie, zhong2025concat}.
However, existing cross-pair datasets are primarily limited to facial domains, making them difficult to generalize to general subject scenarios.
Overall, current training datasets either provide insufficient reference variation or lack domain diversity, limiting the effectiveness of easing copy-paste problem.

In this work, we introduce \textit{Phantom-Data}, a subject-to-video dataset specifically constructed to mitigate the prevalent copy-paste problem in the general scenarios.  It is built around three core design principles for the reference subject: 1) \textbf{General and input aligned subjects}: Reference images should span a wide range of commonly encountered subject types and reflect the distribution of real-world user inputs. 2) \textbf{Different contexts}: Reference subjects appear in varied conditions—such as different backgrounds, viewpoints, or poses—relative to their counterparts in the target video. This encourages the model to generalize identity preservation under distribution shifts and reduces reliance on spurious identity-irrelevant correlations. 3) \textbf{Consistent identity}: Despite contextual variation, the reference subject must remain visually consistent with the target video subject in terms of shape, structure, and texture.

To fulfill these principles, we design a three-stage pipeline:
Firstly, we perform \textbf{S2V Detection} by leveraging a vision-language model to conduct open-set object detection and identify candidate subjects of appropriate size. A second-stage filtering step further refines the results by retaining only subjects that are both semantically relevant and visually compact.
Then, we conduct \textbf{Contextually Diverse Retrieval} by constructing a large-scale subject database comprising over 53 million video segments and 3 billion image samples, increasing the likelihood of retrieving the same identity under diverse backgrounds, poses, and viewpoints.
Finally, we apply \textbf{Prior-Guided Identity Verification} to ensure identity consistency. For living beings (e.g., humans, animals), we mine temporal structures from long videos to construct cross-context pairs. For static objects (e.g., products), we perform category-specific retrieval. A final VLM-based pairwise check verifies that each selected pair maintains both identity consistency and contextual diversity. Through this pipeline, we construct a large-scale, high-quality cross-pair consistency dataset comprising approximately \textit{1 million identity-consistent pairs}  with over \textit{30,000 multi-subject scenes}, offering a strong foundation for modeling general subject-to-video tasks. Representative samples are shown in Fig.~\ref{fig:teaser}.

To validate the effectiveness of our dataset, we conduct comprehensive experiments using open-source video generation models. The results demonstrate that, compared to prior data construction methods, our cross-pair approach substantially improves key metrics such as text alignment and visual quality, while maintaining identity consistency on par with in-pair baselines.
Furthermore, we perform ablation studies to highlight the importance of a large-scale and diverse cross-pair dataset, showing that both data volume and scene diversity play a critical role in enhancing generation performance. We also validate the effectiveness of our data  pipeline in preserving high identity consistency while ensuring sufficient contextual diversity.

Our main contributions can be summarized as follows:
\begin{itemize}
\item We introduce Phantom-Data, the first general-purpose cross-pair video consistency dataset, comprising approximately 1 million high-quality, identity-consistent pairs that span a wide range of subject categories and visual contexts.

\item We present a structured data-construction pipeline purpose-built for subject-consistent video generation. It unifies a subject-centric detection module optimized for the S2V task, large-scale cross-context retrieval, and prior-guided identity verification, thereby securing strict identity fidelity while introducing rich contextual diversity.

\item We conduct extensive experiments to validate the effectiveness of our dataset, demonstrating consistent improvements in text alignment, visual quality, and generalization over existing in-pair baselines.
\end{itemize}

\begin{table}[tbh]
\small 
\centering
\renewcommand{\arraystretch}{1.2}
\begin{tabular}{lccccc}
\hline
Method & General Objects & Input-aligned Objects & Diverse Context  & Publicly Available  \\
\hline
MovieGen\cite{polyak2024movie}       & \xmark & \cmark & 
 \cmark & \xmark  \\
Video Alchemist\cite{chen2025multi} & \cmark & \xmark & \xmark &  \xmark  \\
ConceptMaster\cite{huang2025conceptmaster}  & \cmark & \xmark & \xmark &  \xmark \\
\hline
Ours            & \cmark & \cmark & \cmark &  \cmark \\
\hline
\end{tabular}
\caption{Comparison between Phantom-data and datasets used in prior work.}
\label{tab:dataset_comparison}
\end{table}

\section{Related Work}\label{sec:related}

\myparagraph{Text-to-Video Generation.} Early diffusion-based video generators \cite{blattmann2023stable, guo2024animatediff, wang2024magicvideo} were limited to producing short clips with constrained spatial and temporal resolution. However, the field has rapidly progressed with the introduction of large-scale latent diffusion models and transformer-based architectures. Notably, Sora~\cite{sora} is capable of generating minute-long, high-fidelity videos, while contemporaneous systems such as Seaweed~\cite{seawead2025seaweed}, Hunyuan-Video~\cite{kong2024hunyuanvideo}, CogVideo-X~\cite{yang2024cogvideox}, MAGI~\cite{magi1}, and others~\cite{wan2025} have further advanced frame rate, resolution, scene complexity, realism, and motion smoothness. Despite their impressive visual quality, these generic text-conditioned models provide only coarse control: textual prompts alone cannot fully specify scene layout, subject appearance, or viewpoint, motivating research into finer control signals. 

\myparagraph{Subject-Consistent Video Generation.} The task of subject-consistent video generation (S2V)~\cite{liu2025phantom, fei2025skyreels, chen2025multi, huang2025conceptmaster, deng2025cinema} focuses on generating videos that not only align with the given text prompt but also preserve the visual identity of a reference subject, such as a person, animal, product, or scene.
From a modeling perspective, one common strategy \cite{huang2025conceptmaster, chen2025multi, polyak2024movie, hu2025hunyuancustom} is cross-attention-based fusion, where visual features extracted from pretrained encoders\cite{radford2021learning, oquab2024dinov2, xudemystifying} or VLMs, are injected into the generative backbone through dedicated attention layers. 
An alternative approach is noise-space conditioning, where identity features obtained from a VAE encoder are directly concatenated with the noise input of the diffusion model, without modifying the underlying architecture. This lightweight design enables nearly lossless injection of identity information, as seen in DIT-style models such as Phantom~\cite{liu2025phantom} and VACE~\cite{jiang2025vace}.
Recent systems like SkyReels-A2~\cite{fei2025skyreels} explore combining both strategies, incorporating cross-attention guidance and noise-level conditioning within a unified framework.

\myparagraph{Training Data in Subject-to-Video Generation.} Training data plays a crucial role in subject-consistent video generation, as it directly influences a model’s ability to generate faithful and controllable results. Most existing approaches rely on \textit{in-pair} supervision, where the reference and target frames are sampled from the same video clip. While this setup guarantees identity alignment, it often leads to the undesirable \textit{copy-paste} effect—where the model reproduces not only the subject but also the background and pose of the reference frame, limiting its capacity to follow the input prompt.
To mitigate this issue, several works~\cite{chen2025multi, liang2025movie, huang2025conceptmaster, jiang2025vace, ju2025fulldit} adopt data normalization and augmentation strategies, such as background removal, color jittering, and geometric transformations. However, these techniques, combined with the limited diversity inherent in in-pair training, are often insufficient to address complex contextual variations such as motion, viewpoint, and scene layout. Recent efforts have turned to \textit{cross-pair} training, where identity-consistent reference and target frames are sampled from different videos. This setting encourages the model to concentrate on subject identity while reducing overfitting to specific visual contexts~\cite{polyak2024movie, zhong2025concat}. Nevertheless, current cross-pair datasets are mostly restricted to narrow domains like human faces, limiting their generalizability to broader subject categories such as animals, products, or stylized characters. In summary, although cross-pair supervision offers a promising direction for addressing the copy-paste issue, the absence of high-quality, diverse, and identity-consistent training data across general domains remains a significant bottleneck for advancing  S2V models. To bridge this gap, we introduce Phantom-data, a large-scale cross-pair dataset designed to support subject-consistent video generation across a wide range of real-world categories.


\section{Phantom Data}\label{sec:intro}
\begin{figure}[ht]
  \centering
  \includegraphics[width=\linewidth]{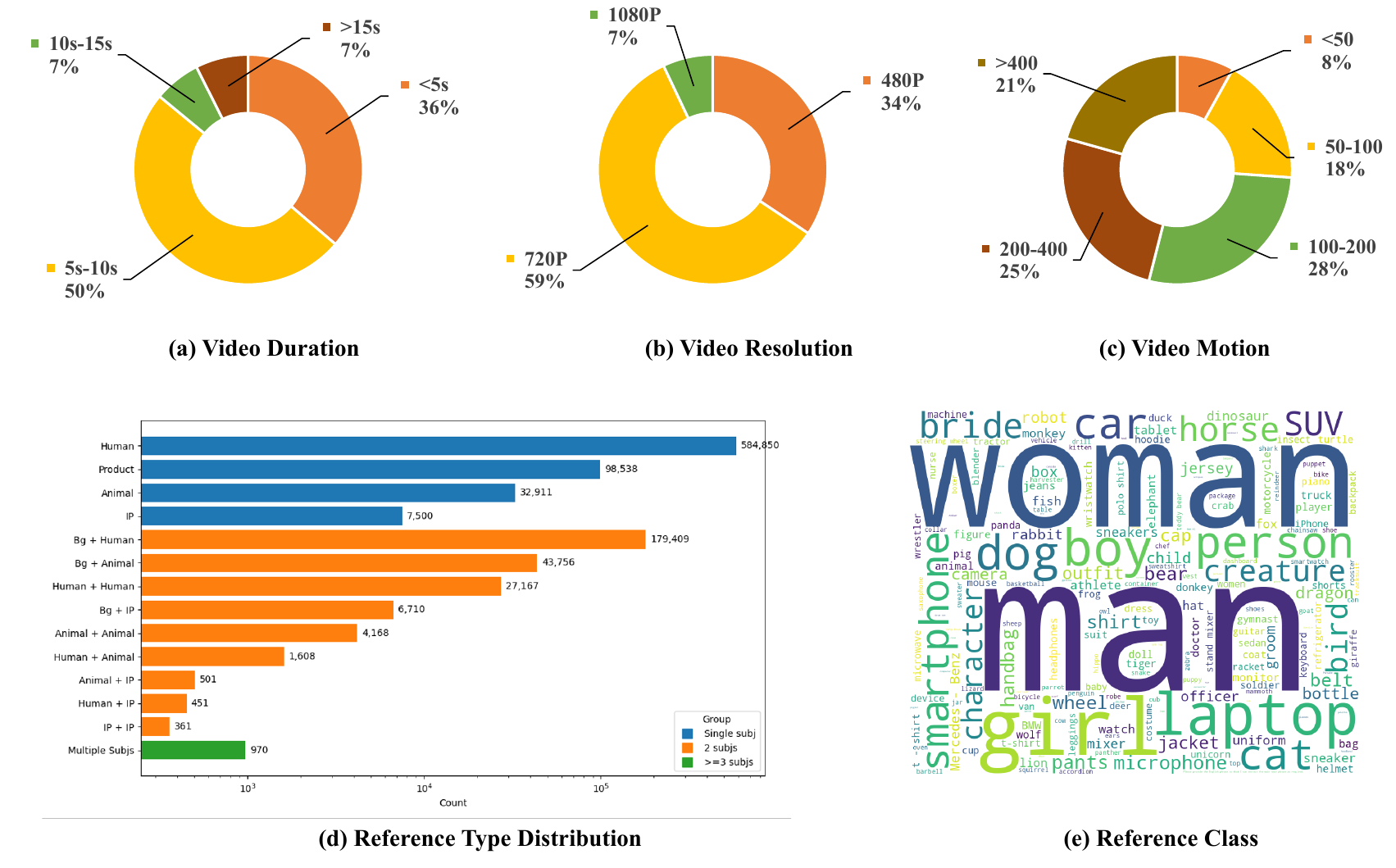}
  \caption{The statistical analysis of Phantom-Data.}
  \label{fig:dataset_overview}
\end{figure}

We provide a detailed analysis of \textbf{Phantom-Data}, focusing on its statistical properties and a comparison with existing datasets for subject-consistent video generation.

\subsection{Statistical Analysis} 

We analyze the dataset at both the video and subject levels.

\myparagraph{Video-level properties.}  
As shown in Fig.~\ref{fig:dataset_overview}(a–c), our dataset spans a wide range of video durations, resolutions, and motion patterns. Around 50\% of videos are 5--10 seconds long, and the majority are in 720p resolution. Motion levels also vary considerably, covering both relatively static and highly dynamic scenes.

\myparagraph{Subject composition.}  
Fig.~\ref{fig:dataset_overview}(d) illustrates the distribution of subject types and their combinations. While the majority of samples (approximately 720{,}000) contain a single subject—such as a human, product, or animal—a substantial portion (around 280{,}000) involve two or more co-occurring entities, supporting multi-subject consistency modeling.

\myparagraph{Reference diversity.} As shown in Fig.~\ref{fig:dataset_overview}(e), the dataset spans a broad semantic space of subject categories. Common reference entities include humans (e.g., woman, man, girl), animals (e.g., dog, bird), and man-made objects (e.g., smartphone, car, laptop), highlighting the dataset's suitability for general-purpose subject-to-video modeling across varied domains.

\subsection{Comparison with Prior Datasets}  
As summarized in Table~\ref{tab:dataset_comparison}, existing datasets for subject-consistent video generation either lack general object coverage, rely heavily on input-aligned references from the same video, or are limited in contextual diversity. In contrast, Phantom-Data offers a more comprehensive setting: it supports general object categories beyond faces, encourages cross-context modeling by sampling subject-reference pairs from diverse scenes, and is publicly available for research. This makes it the first open-access dataset to jointly support identity consistency and context diversity in a general-purpose, cross-pair setup.
\section{Data Pipeline}\label{sec:method}

\subsection{Video Data Source}
The Phantom-Data video dataset consists of clips collected from public sources such as Koala-36M~\cite{wang2024koala}, as well as proprietary internal repositories. Each video undergoes a rigorous quality control pipeline, including black border detection, motion analysis, and other filtering steps. Subsequently, long videos are segmented into short clips at the second level using scene segmentation. Each resulting clip is then annotated with a corresponding video caption. The total number of videos is approximately 53 million.

\begin{figure}[!tbh]
  \centering
  \includegraphics[width=\linewidth]{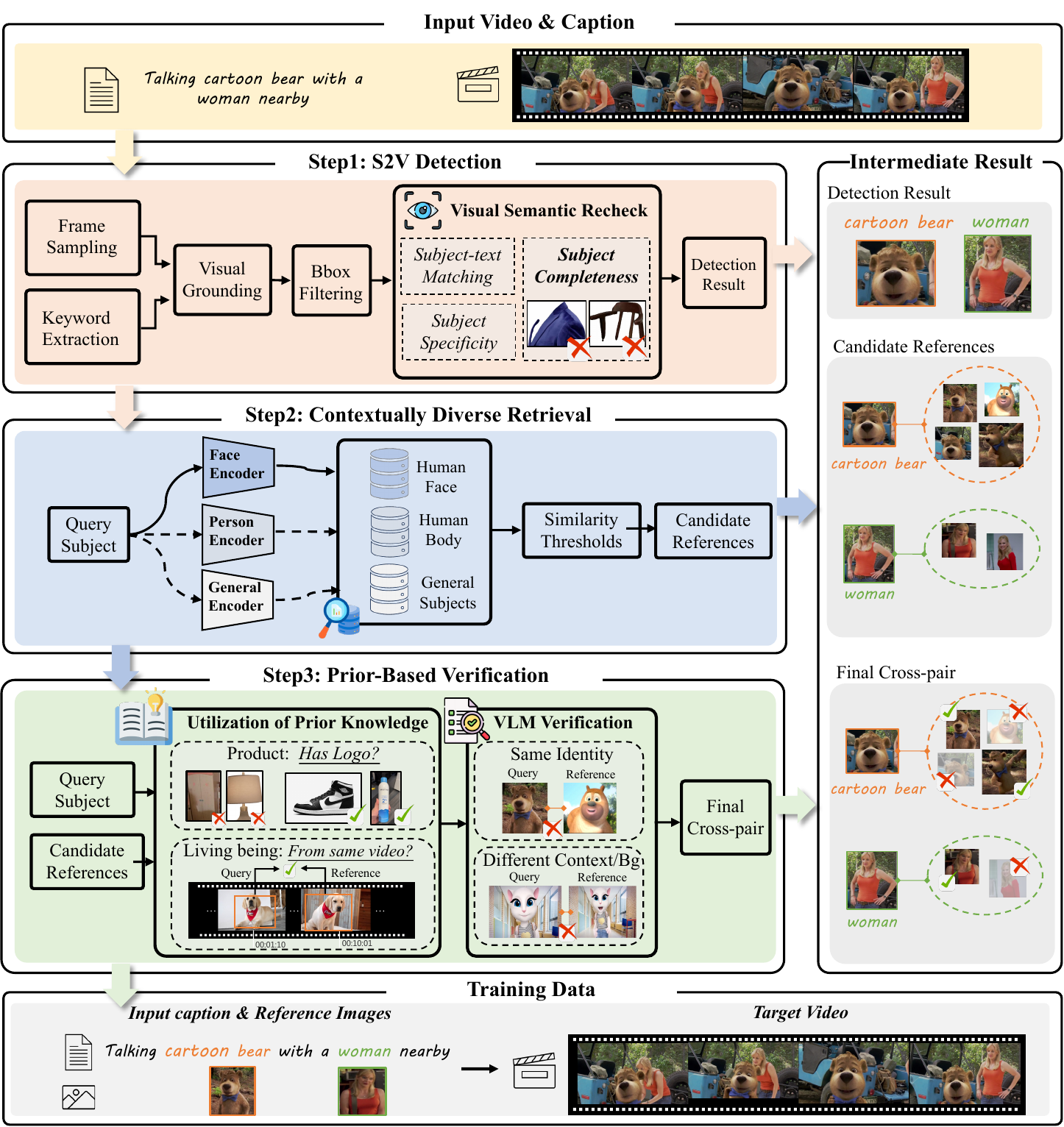}
  \caption{The overview of the data pipeline for constructing cross-pair training samples. 
  }
  \label{fig:pipeline}
\end{figure}

\subsection{Data Pipeline}

Given an input video and its associated caption, we focus on constructing a high-quality \textit{cross-pair} dataset, where the same subject appears across different visual contexts while maintaining identity consistency. To this end, we design a structured data pipeline consisting of three key stages. As shown in Fig.\ref{fig:pipeline}, firstly, we perform \textit{S2V Detection} to identify high-quality subject instances from videos. Then, we propose a \textit{Contextually Diverse Retrieval} module to recall candidate images that are likely to correspond to the detected subjects across varying scenes. Finally, we apply \textit{Prior-based Identity Verification} to filter the retrieved candidates, ensuring that only those sharing the same identity across different contexts are retained.







\subsubsection{S2V Detection}

This stage aims to identify diverse and qualified subjects from each video clip as candidates for cross-scene pairing. It consists of five major steps:





\myparagraph{1. Frame Sampling.}
To reduce computation, we sample three frames at $t = 0.05$, $0.5$, and $0.95$ of each clip, following \cite{chen2025multi}, ensuring temporal diversity while avoiding full-frame processing.

\myparagraph{2. Keyword Extraction.}
We use Qwen2.5 \cite{yang2024qwen2} to extract key noun phrases (e.g., people, animals, products) from captions, serving as subject candidates for grounding.

\myparagraph{3. Visual Grounding.}
Qwen2.5-VL \cite{bai2025qwen2} aligns each phrase to regions in the sampled frames. Ambiguous matches mapping to multiple regions are removed to reduce noise.

\myparagraph{4. Bbox Filtering.}
We retain boxes covering between $4\%$ and $90\%$ of the image and at least $128 \times 128$ in size. Overlapping boxes (IoU > 0.8) are suppressed for clarity.



 \myparagraph{5. Visual-Semantic Recheck.}  
To further ensure the quality of the grounded subjects, we employ another vision-language model, InternVL2.5 7B \cite{chen2024expanding}, to validate each detection against the following criteria:  1) Completeness: We observe that visual grounding often produces bounding boxes around partial or cropped objects, as a result of the underlying detection model's exhaustive labeling strategy. However, for S2V task, users typically provide complete reference subjects, making such incomplete detections unsuitable. We therefore filter out any region that fails to cover the full extent of the object.
    2) Specificity: The subject must be visually distinct and identifiable. Vague or generic objects, such as trees, rocks, or background clutter, are excluded.
  3) Subject-text 
Matching: The grounded region must be semantically consistent with the associated phrase. To improve alignment precision, we employ a separate instance of InternVL2.5 to reevaluate the consistency between the textual description and the detected subject.


As a result of this pipeline, we obtain a high-quality set of subject instances, each paired with a corresponding descriptive phrase. Since a subject may appear in multiple frames across the video, we select only one representative instance for visualization in the Intermediate Result section of Fig.~\ref{fig:pipeline}.

\subsubsection{Contextually Diverse Retrieval}

Given the subject instances detected in the previous stage, we aim to find candidate reference images of the same subject appearing in different visual contexts. To achieve this, we construct a large-scale retrieval bank and use the detected subjects to perform identity-aware querying.

\myparagraph{Large-Scale Retrieval Bank Construction.}  
The retrieval bank comprises two essential components: diverse subject image sources to increase contextual variability, and feature representations tailored for identity-preserving retrieval.

\textit{Subject Source.}  
We begin by registering every detected subject instance from the training videos into a retrieval bank. To further broaden candidate diversity, we augment this bank with an extra 3 billion images from the LAION dataset \cite{schuhmann2022laion} beyond the original video corpus. These external images inject greater variation in scene, pose, and appearance, delivering broader contextual coverage during retrieval, an advantage that is particularly valuable for product-centric scenarios with substantial intra-instance variation.

\textit{Subject Representation.}  
To support reliable cross-context identity matching, we employ expert-designed encoders to extract identity-preserving and context-invariant embeddings tailored to different subject categories. These embeddings are used for both indexing the retrieval bank and querying.

For facial representation, we adopt the widely used ArcFace encoder~\cite{deng2019arcface} to extract robust and discriminative identity embeddings:
\begin{equation}
    V_{\text{face}} = E_{\text{arcface}}(I_{\text{face}}).
\end{equation}

For general objects, inspired by ObjectMate~\cite{winter2024objectmate}, we employ a CLIP-based model fine-tuned on a consistency-focused image dataset~\cite{link1} to extract identity-preserving embeddings:
\begin{equation}
    V_{\text{subj}} = E_{\text{IR}}(I).
\end{equation}

For human subjects, which are central to many downstream applications, we combine both facial and clothing features. Each individual is represented by concatenating the general appearance embedding with the corresponding facial embedding:
\begin{equation}
    V_{\text{person}} = [E_{\text{IR}}(I), E_{\text{arcface}}(I_{\text{face}})].
\end{equation}



\myparagraph{Query-Based Retrieval.} To ensure the retrieved candidates are visually distinct from the query image yet share the same identity, we apply both upper and lower bounds on similarity. Specifically, we discard overly similar results (potential duplicates) by enforcing an upper similarity threshold, and exclude unrelated identities by applying a lower threshold.

\subsubsection{Prior-Based Identity Verification}

However, due to the large scale of the retrieval corpus, false positives 
frequently occur even within seemingly reasonable similarity ranges. To address this issue, we adopt a two-stage filtering strategy based on prior knowledge and VLM Verification.

\myparagraph{Utilization of Prior Knowledge.}  
We apply category-specific filtering strategies to improve cross-pair reliability: 1) \textit{Non-living subjects} (e.g., products): These typically exhibit high intra-class variability, making identity verification more challenging. To improve precision, we retain only product instances that feature complete and recognizable brand logos (e.g., Nike, Audi), which remain visible across different scenes. 2) \textit{Living entities} (e.g., humans, animals): For these subjects, we restrict retrieved candidates to those from different clips within the same long-form video. This constraint ensures natural variation in scene and pose while maintaining consistent identity.

\myparagraph{VLM-Based Consistency Verification.}  
To further ensure both identity consistency and contextual diversity, we apply a VLM-based verification procedure: 1) \textit{identity consistency}:  For non-living objects, we enforce strict similarity in visual details such as color, packaging, and textual elements, while allowing for background variation. For living subjects, especially humans, we verify facial identity consistency and, in the case of full-body samples, also ensure clothing alignment. 2) \textit{Contextual diversity}. We keep only those cross-pair samples that exhibit substantial variation in background and scene context, thereby alleviating copy-paste artifacts during model training.

 

\section{Experiments}\label{sec:exp}

\subsection{Implementation}

\myparagraph{Model Architecture.}
We validate the effectiveness of our proposed data using the Phantom-wan \cite{liu2025phantom} model. Built on the Wan2.1 \cite{wang2025wan} foundation, Phantom-wan is a leading open-source framework for subject-consistent video generation. 

\myparagraph{Training and inference.}
We train a 1.3 billion-parameter Phantom-wan model using Rectified Flow (RF) \cite{lipman2022flow, liuflow} as the training objective. The training is performed on 64 A100 GPUs for 30k iterations with 480p resolution data, which yields stable performance. During inference, we apply Euler sampling with 50 steps and use classifier-free guidance \cite{ho2022classifier} to decouple image and text conditions. All experiments follow the same training and inference settings to ensure fair comparisons.

\myparagraph{Evaluation.} We construct a test suite of 100 cases from diverse scenarios, covering humans, animals, products, environments, and clothing. These cases include both single- and multi-subject settings, paired with manually written text prompts that reflect natural user input. 

We evaluate model performance across three dimensions: video quality, text-video consistency, and subject-video consistency. Subject-video consistency is evaluated using CLIP \cite{guo2024pulid}, DINO \cite{oquab2023dinov2}, and GPT-4o scores, following recent evaluation protocols inspired by \cite{peng2024dreambench++}. Text-video consistency is measured via Reward-TA \cite{liu2025improving}, both of which assess the semantic alignment between generated video content and the text prompt. Video quality is assessed using VBench \cite{huang2024vbench} , which provide fine-grained evaluation along several aspects, including Temporal (temporal flickering and stability), Motion (smoothness of subject motion), IQ (overall imaging quality), BG (background consistency across frames), and Subj (temporal consistency of the generated subject).

\subsection{Main Results}
\begin{figure}[t]
  \centering
  \includegraphics[width=\linewidth]{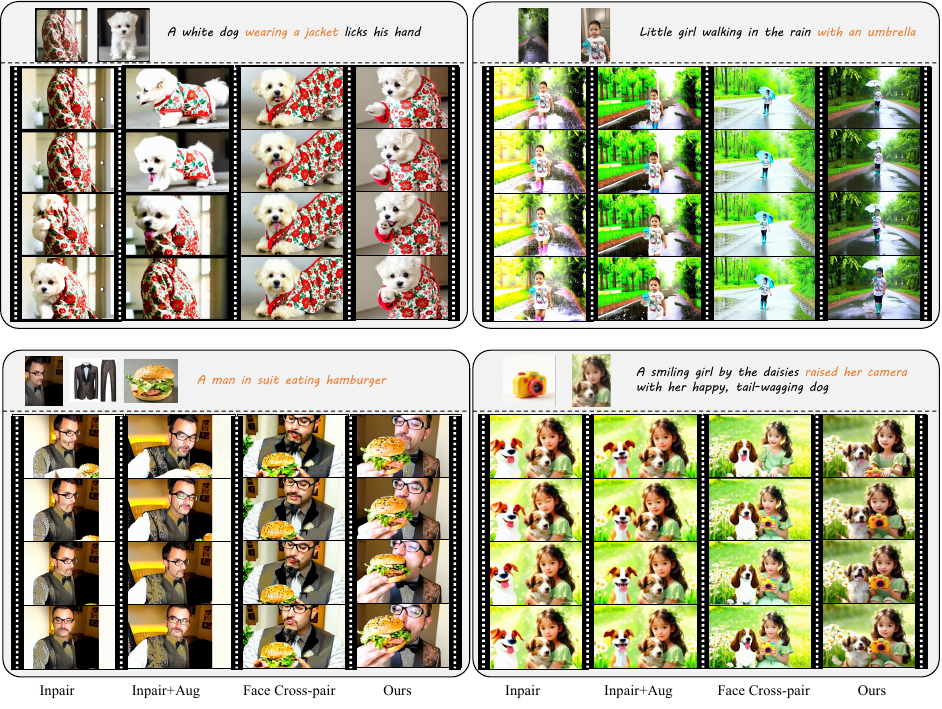}
  \caption{Qualitative comparisons across different training strategies. 
  }
  \label{fig:main_result}
\end{figure}

\begin{table}[t]
    \centering
    \resizebox{\columnwidth}{!}{
    \small
    \begin{tabular}{lcccccccc}
        \toprule
        \multirow{2}{*}{\vtop{\hbox{\strut Methods}}} 
        & \multicolumn{2}{c}{Subject Consistency} 
        & \multicolumn{1}{c}{Prompt Following} 
        & \multicolumn{5}{c}{Video Quality} \\
        \cmidrule(lr){2-3} \cmidrule(lr){4-4} \cmidrule(lr){5-9}
        & DINO $\uparrow$ & GPT-4o $\uparrow$ & Reward-TA $\uparrow$ 
        & Temporal $\uparrow$ & Motion $\uparrow$ & IQ $\uparrow$ & BG $\uparrow$ & Subj $\uparrow$ \\
        \midrule
        In-pair 
        & \textbf{0.478} & 2.481 & 2.074 
        & 0.971 & 0.985 & 0.725 & 0.937 & 0.933 \\
        
        In-pair + Data Aug 
        & \underline{0.473} & \underline{2.792} & 2.427 
        & 0.961 & 0.979 & \underline{0.730} & 0.932 & 0.922 \\
        
        Face Cross-pair 
        & 0.354 & 2.378 & \underline{3.022}
        & \textbf{0.983} & \textbf{0.989} & 0.723 & \underline{0.937} & \underline{0.935} \\
        
        Ours 
        & 0.416 & \textbf{3.041} & \textbf{3.827} 
        & \underline{0.975} & \underline{0.986} & \textbf{0.739} & \textbf{0.948} & \textbf{0.944} \\
        \bottomrule
    \end{tabular}
    }
    \caption{Main results comparing prompt following, subject consistency, and video quality across different training paradigms. Bold denotes the best performance per column. The \underline{underline} indicates the second-highest scores.}
    \label{tab:main_res}
\end{table}

We evaluate our method against three representative baselines: (1) In-pair training, which samples the reference subject from the same video; (2) In-pair with copy-augmentation, which introduces spatial and appearance augmentations to reduce overfitting as in \cite{chen2025multi}; and (3) Face-based cross-pair, which utilizes face-level identity matching across videos. We report both quantitative and qualitative comparisons.

\begin{figure}[!t]
  \centering
  \includegraphics[width=0.95\linewidth]{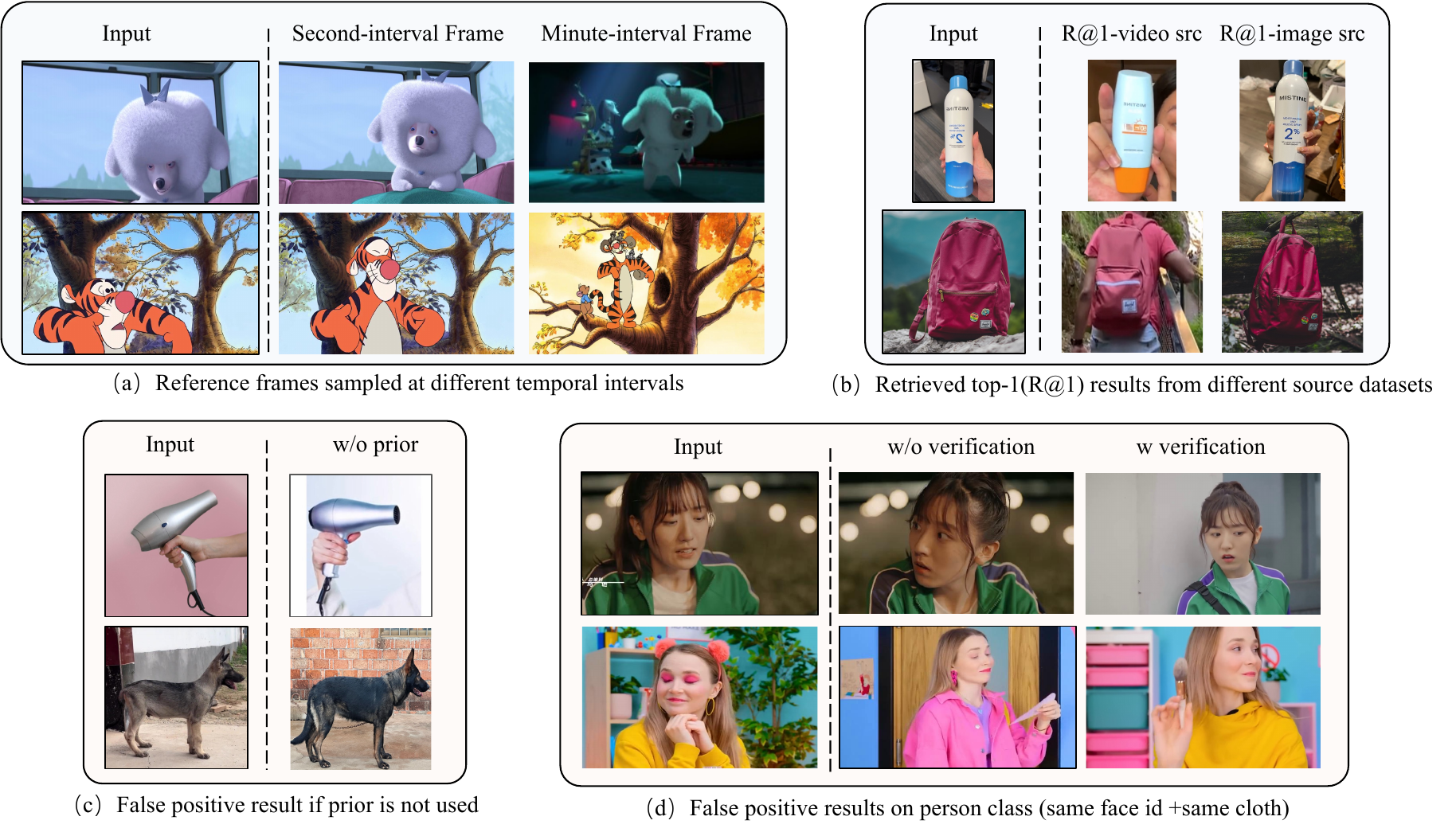}

\caption{Ablation study on Contextually Diverse Retrieval and Prior-Based Identity Verification. 
(a) Reference frames from different timestamps show that longer videos offer more diverse contexts. 
(b) Retrieval from large-scale image datasets improves recall and candidate diversity. 
(c) Without prior filtering, false positives may be included. 
(d) Verification removes mismatched or overly similar identities, ensuring high-quality pairs.}
  \label{fig:ablation_study}
\end{figure}
\begin{figure}[thb]
\CenterFloatBoxes
\begin{floatrow}

\ttabbox
{
\resizebox{0.95\linewidth}{!}{
\begin{tabular}{lccc}
    \toprule
    \multirow{2}{*}{\vtop{\hbox{\strut Methods}}} 
    & \multicolumn{2}{c}{Subject Consistency} 
    & \multicolumn{1}{c}{Prompt Following} \\
    \cmidrule(lr){2-3} \cmidrule(lr){4-4}
    & DINO $\uparrow$ & GPT-4o $\uparrow$ & reward-TA $\uparrow$ \\
    \midrule
    baseline (face only)   
        & 0.354 & 2.378 & 3.022 \\
    + human                
        & 0.401\textsubscript{{\small +0.047}} 
        & 2.747\textsubscript{{\small +0.363}} 
        & 3.726\textsubscript{{\small +0.702}} \\
    + IP/animal            
        & 0.416\textsubscript{{\small +0.062}} 
        & 2.795\textsubscript{{\small +0.411}} 
        & 3.407\textsubscript{{\small +0.383}} \\
    + product              
        & 0.386\textsubscript{{\small +0.032}} 
        & 2.662\textsubscript{{\small +0.288}} 
        & 3.572\textsubscript{{\small +0.58}} \\
    + multi-subject        
        & 0.418\textsubscript{{\small +0.064}} 
        & 2.901\textsubscript{{\small +0.525}} 
        & 3.512\textsubscript{{\small +0.498}} \\
    \bottomrule
\end{tabular}}}
{\caption{Ablation study on subject diversity. }
 \label{tab:abl_subject}
}

\ttabbox
{
\resizebox{0.95\linewidth}{!}{
\begin{tabular}{lccc}
    \toprule
    \multirow{2}{*}{\vtop{\hbox{\strut Methods}}} 
    & \multicolumn{2}{c}{Subject Consistency} 
    & \multicolumn{1}{c}{Prompt Following} \\
    \cmidrule(lr){2-3} \cmidrule(lr){4-4}
    & DINO $\uparrow$ & GPT-4o $\uparrow$ & reward-TA $\uparrow$ \\
    \midrule
    100k   & 0.408 & 3.090 & 3.796 \\
    1 M  & \textbf{0.416} & \textbf{3.175} & \textbf{3.827} \\
    \bottomrule
\end{tabular}}}
{\caption{Ablation study on data scale.}
 \label{tab:abl_scale}
}

\end{floatrow}
\end{figure}

Quantitative results demonstrate that our cross-pair training paradigm achieves state-of-the-art performance in terms of text-video alignment and overall video quality, as measured by reward-based evaluation metrics. Furthermore, our method delivers competitive subject consistency, rivaling in-pair baselines despite the increased scene diversity. In contrast, In-pair settings suffer from poor text-following ability due to overfitting on narrow visual contexts. The Face cross-pair method performs slightly better on prompt following but is limited by the narrow domain of its face-centric training data, resulting in weaker identity preservation across diverse subjects.

Qualitative comparisons, as shown in Fig.\ref{fig:main_result}, further support our conclusions. Across multiple prompts and subject categories, models trained with in-pair data consistently fail to follow textual instructions, often generating videos with obvious artifacts. In contrast, our cross-pair trained model successfully aligns with the prompt across all cases, producing coherent and faithful subject-driven videos.

\subsection{Ablation Studies}
\myparagraph{Subject Diversity.}   As shown in Table~\ref{tab:abl_subject}, enriching the training set with diverse subject types—including humans, animals, products, and multi-subject scenes—consistently improves subject consistency and prompt following, compared to the face-only baseline.

\myparagraph{Data Scale.} Table~\ref{tab:abl_scale} illustrates the effect of data scale. Increasing the training set from 100 k to 1 million samples leads to further improvements across all metrics, highlighting the importance of both diversity and scale in building a robust subject-to-video generation dataset.

\myparagraph{Contextually Diverse Retrieval.}
To assess the impact of contextual diversity in reference selection, we compare different sampling and retrieval strategies: (1) Temporal Sampling. As illustrated in Fig.~\ref{fig:ablation_study}(a), reference frames sampled at longer temporal intervals (e.g., minute-level vs. second-level) provides richer visual diversity. (2) Multi-source Retrieval. Fig.~\ref{fig:ablation_study}(b) compares retrieval from video-only sources and from a combined image+video retrieval bank. Incorporating large-scale image datasets improves both recall and candidate diversity.

\myparagraph{Prior-Based Identity Verification.}
We further evaluate the role of prior filtering and identity verification in ensuring training quality: (1) Prior Filtering. Without prior-based constraints, visually similar but semantically incorrect matches (false positives) are often included (see Fig.~\ref{fig:ablation_study}(c)). (2) Verification Module. As shown in Fig.~\ref{fig:ablation_study}(d), our identity verification module further refines the candidate set by removing both near-duplicates (overly similar samples) and mismatched identities (overly dissimilar ones).

\section{Conclusion}\label{sec:con}
We propose \textbf{Phantom-Data}, a large-scale, general-purpose cross-pair dataset to improve subject consistency and text alignment in text-to-video generation. By introducing a structured pipeline—combining open-vocabulary detection, diverse cross-context retrieval, and identity verification—we address the limitations of in-pair training and reduce the copy-paste problem. Experiments show that our dataset significantly boosts generation quality while maintaining strong identity consistency. Phantom-Data provides a solid foundation for future research in controllable subject-to-video generation task.

\clearpage

\bibliographystyle{plainnat}
\bibliography{main}

\begin{thebibliography}{50}
\providecommand{\natexlab}[1]{#1}
\providecommand{\url}[1]{\texttt{#1}}
\expandafter\ifx\csname urlstyle\endcsname\relax
  \providecommand{\doi}[1]{doi: #1}\else
  \providecommand{\doi}{doi: \begingroup \urlstyle{rm}\Url}\fi

\bibitem[Kli()]{Kling}
Kling.
\newblock \url{https://app.klingai.com/cn/image-to-video/frame-mode/new}.

\bibitem[Bai et~al.(2025)Bai, Chen, Liu, Wang, Ge, Song, Dang, Wang, Wang, Tang, et~al.]{bai2025qwen2}
Shuai Bai, Keqin Chen, Xuejing Liu, Jialin Wang, Wenbin Ge, Sibo Song, Kai Dang, Peng Wang, Shijie Wang, Jun Tang, et~al.
\newblock Qwen2. 5-vl technical report.
\newblock \emph{arXiv preprint arXiv:2502.13923}, 2025.

\bibitem[Blattmann et~al.(2023)Blattmann, Dockhorn, Kulal, Mendelevitch, Kilian, Lorenz, Levi, English, Voleti, Letts, et~al.]{blattmann2023stable}
Andreas Blattmann, Tim Dockhorn, Sumith Kulal, Daniel Mendelevitch, Maciej Kilian, Dominik Lorenz, Yam Levi, Zion English, Vikram Voleti, Adam Letts, et~al.
\newblock Stable video diffusion: Scaling latent video diffusion models to large datasets.
\newblock \emph{arXiv preprint arXiv:2311.15127}, 2023.

\bibitem[Brooks et~al.(2024)Brooks, Peebles, Holmes, DePue, Guo, Jing, Schnurr, Taylor, Luhman, Luhman, Ng, Wang, and Ramesh]{sora}
Tim Brooks, Bill Peebles, Connor Holmes, Will DePue, Yufei Guo, Li~Jing, David Schnurr, Joe Taylor, Troy Luhman, Eric Luhman, Clarence Ng, Ricky Wang, and Aditya Ramesh.
\newblock Video generation models as world simulators, 2024.
\newblock URL \url{https://openai.com/research/video-generation-models-as-world-simulators}.

\bibitem[Chen et~al.(2025{\natexlab{a}})Chen, Ge, Zhang, Zhang, Zhu, Yang, Hao, Wu, Lai, Hu, et~al.]{chen2025goku}
Shoufa Chen, Chongjian Ge, Yuqi Zhang, Yida Zhang, Fengda Zhu, Hao Yang, Hongxiang Hao, Hui Wu, Zhichao Lai, Yifei Hu, et~al.
\newblock Goku: Flow based video generative foundation models.
\newblock \emph{arXiv preprint arXiv:2502.04896}, 2025{\natexlab{a}}.

\bibitem[Chen et~al.(2025{\natexlab{b}})Chen, Siarohin, Menapace, Fang, Lee, Skorokhodov, Aberman, Zhu, Yang, and Tulyakov]{chen2025multi}
Tsai-Shien Chen, Aliaksandr Siarohin, Willi Menapace, Yuwei Fang, Kwot~Sin Lee, Ivan Skorokhodov, Kfir Aberman, Jun-Yan Zhu, Ming-Hsuan Yang, and Sergey Tulyakov.
\newblock Multi-subject open-set personalization in video generation.
\newblock \emph{arXiv preprint arXiv:2501.06187}, 2025{\natexlab{b}}.

\bibitem[Chen et~al.(2024)Chen, Wang, Cao, Liu, Gao, Cui, Zhu, Ye, Tian, Liu, et~al.]{chen2024expanding}
Zhe Chen, Weiyun Wang, Yue Cao, Yangzhou Liu, Zhangwei Gao, Erfei Cui, Jinguo Zhu, Shenglong Ye, Hao Tian, Zhaoyang Liu, et~al.
\newblock Expanding performance boundaries of open-source multimodal models with model, data, and test-time scaling.
\newblock \emph{arXiv preprint arXiv:2412.05271}, 2024.

\bibitem[Deng et~al.(2019)Deng, Guo, Xue, and Zafeiriou]{deng2019arcface}
Jiankang Deng, Jia Guo, Niannan Xue, and Stefanos Zafeiriou.
\newblock Arcface: Additive angular margin loss for deep face recognition.
\newblock In \emph{Proceedings of the IEEE/CVF conference on computer vision and pattern recognition}, pages 4690--4699, 2019.

\bibitem[Deng et~al.(2025)Deng, Guo, Wang, Fang, Wang, Yuan, Yang, Liu, Huang, and Ma]{deng2025cinema}
Yufan Deng, Xun Guo, Yizhi Wang, Jacob~Zhiyuan Fang, Angtian Wang, Shenghai Yuan, Yiding Yang, Bo~Liu, Haibin Huang, and Chongyang Ma.
\newblock Cinema: Coherent multi-subject video generation via mllm-based guidance.
\newblock \emph{arXiv preprint arXiv:2503.10391}, 2025.

\bibitem[Fei et~al.(2025)Fei, Li, Qiu, Wang, Dou, Wang, Xu, Fan, Chen, Li, et~al.]{fei2025skyreels}
Zhengcong Fei, Debang Li, Di~Qiu, Jiahua Wang, Yikun Dou, Rui Wang, Jingtao Xu, Mingyuan Fan, Guibin Chen, Yang Li, et~al.
\newblock Skyreels-a2: Compose anything in video diffusion transformers.
\newblock \emph{arXiv preprint arXiv:2504.02436}, 2025.

\bibitem[Feng et~al.(2025)Feng, Liu, Tu, Qi, Sun, Ma, Zhao, Zhou, and He]{i2vcontrolcamera}
Wanquan Feng, Jiawei Liu, Pengqi Tu, Tianhao Qi, Mingzhen Sun, Tianxiang Ma, Songtao Zhao, Siyu Zhou, and Qian He.
\newblock I2vcontrol-camera: Precise video camera control with adjustable motion strength.
\newblock 2025.

\bibitem[Guo et~al.(2024{\natexlab{a}})Guo, Yang, Rao, Liang, Wang, Qiao, Agrawala, Lin, and Dai]{guo2024animatediff}
Yuwei Guo, Ceyuan Yang, Anyi Rao, Zhengyang Liang, Yaohui Wang, Yu~Qiao, Maneesh Agrawala, Dahua Lin, and Bo~Dai.
\newblock Animatediff: Animate your personalized text-to-image diffusion models without specific tuning.
\newblock In \emph{12th International Conference on Learning Representations, ICLR 2024}, 2024{\natexlab{a}}.

\bibitem[Guo et~al.(2024{\natexlab{b}})Guo, Wu, Chen, Chen, and He]{guo2024pulid}
Zinan Guo, Yanze Wu, Zhuowei Chen, Lang Chen, and Qian He.
\newblock Pulid: Pure and lightning id customization via contrastive alignment.
\newblock \emph{arXiv preprint arXiv:2404.16022}, 2024{\natexlab{b}}.

\bibitem[Ho and Salimans(2022)]{ho2022classifier}
Jonathan Ho and Tim Salimans.
\newblock Classifier-free diffusion guidance.
\newblock \emph{arXiv preprint arXiv:2207.12598}, 2022.

\bibitem[Ho et~al.(2020)Ho, Jain, and Abbeel]{ho2020denoising}
Jonathan Ho, Ajay Jain, and Pieter Abbeel.
\newblock Denoising diffusion probabilistic models.
\newblock \emph{Advances in neural information processing systems}, 33:\penalty0 6840--6851, 2020.

\bibitem[Hu(2024)]{hu2024animate}
Li~Hu.
\newblock Animate anyone: Consistent and controllable image-to-video synthesis for character animation.
\newblock In \emph{Proceedings of the IEEE/CVF Conference on Computer Vision and Pattern Recognition}, pages 8153--8163, 2024.

\bibitem[Hu et~al.(2025)Hu, Yu, Zhou, Liang, Zhou, Lin, and Lu]{hu2025hunyuancustom}
Teng Hu, Zhentao Yu, Zhengguang Zhou, Sen Liang, Yuan Zhou, Qin Lin, and Qinglin Lu.
\newblock Hunyuancustom: A multimodal-driven architecture for customized video generation.
\newblock \emph{arXiv preprint arXiv:2505.04512}, 2025.

\bibitem[Huang et~al.(2025)Huang, Yuan, Liu, Wang, Wang, Zhang, Wan, Zhang, and Gai]{huang2025conceptmaster}
Yuzhou Huang, Ziyang Yuan, Quande Liu, Qiulin Wang, Xintao Wang, Ruimao Zhang, Pengfei Wan, Di~Zhang, and Kun Gai.
\newblock Conceptmaster: Multi-concept video customization on diffusion transformer models without test-time tuning.
\newblock \emph{arXiv preprint arXiv:2501.04698}, 2025.

\bibitem[Huang et~al.(2024)Huang, He, Yu, Zhang, Si, Jiang, Zhang, Wu, Jin, Chanpaisit, et~al.]{huang2024vbench}
Ziqi Huang, Yinan He, Jiashuo Yu, Fan Zhang, Chenyang Si, Yuming Jiang, Yuanhan Zhang, Tianxing Wu, Qingyang Jin, Nattapol Chanpaisit, et~al.
\newblock Vbench: Comprehensive benchmark suite for video generative models.
\newblock In \emph{Proceedings of the IEEE/CVF Conference on Computer Vision and Pattern Recognition}, pages 21807--21818, 2024.

\bibitem[Jiang et~al.(2025)Jiang, Han, Mao, Zhang, Pan, and Liu]{jiang2025vace}
Zeyinzi Jiang, Zhen Han, Chaojie Mao, Jingfeng Zhang, Yulin Pan, and Yu~Liu.
\newblock Vace: All-in-one video creation and editing.
\newblock \emph{arXiv preprint arXiv:2503.07598}, 2025.

\bibitem[Ju et~al.(2025)Ju, Ye, Liu, Wang, Wang, Wan, Zhang, Gai, and Xu]{ju2025fulldit}
Xuan Ju, Weicai Ye, Quande Liu, Qiulin Wang, Xintao Wang, Pengfei Wan, Di~Zhang, Kun Gai, and Qiang Xu.
\newblock Fulldit: Multi-task video generative foundation model with full attention.
\newblock \emph{arXiv preprint arXiv:2503.19907}, 2025.

\bibitem[Kondratyuk et~al.(2024)Kondratyuk, Yu, Gu, Lezama, Huang, Schindler, Hornung, Birodkar, Yan, Chiu, et~al.]{kondratyuk2024videopoet}
Dan Kondratyuk, Lijun Yu, Xiuye Gu, Jos{\'e} Lezama, Jonathan Huang, Grant Schindler, Rachel Hornung, Vighnesh Birodkar, Jimmy Yan, Ming-Chang Chiu, et~al.
\newblock Videopoet: a large language model for zero-shot video generation.
\newblock In \emph{Proceedings of the 41st International Conference on Machine Learning}, pages 25105--25124, 2024.

\bibitem[Kong et~al.(2024)Kong, Tian, Zhang, Min, Dai, Zhou, Xiong, Li, Wu, Zhang, et~al.]{kong2024hunyuanvideo}
Weijie Kong, Qi~Tian, Zijian Zhang, Rox Min, Zuozhuo Dai, Jin Zhou, Jiangfeng Xiong, Xin Li, Bo~Wu, Jianwei Zhang, et~al.
\newblock Hunyuanvideo: A systematic framework for large video generative models.
\newblock \emph{arXiv preprint arXiv:2412.03603}, 2024.

\bibitem[Liang et~al.(2025)Liang, Ma, He, Hou, Hou, Li, Dai, Juefei-Xu, Azadi, Sinha, et~al.]{liang2025movie}
Feng Liang, Haoyu Ma, Zecheng He, Tingbo Hou, Ji~Hou, Kunpeng Li, Xiaoliang Dai, Felix Juefei-Xu, Samaneh Azadi, Animesh Sinha, et~al.
\newblock Movie weaver: Tuning-free multi-concept video personalization with anchored prompts.
\newblock \emph{arXiv preprint arXiv:2502.07802}, 2025.

\bibitem[Lipman et~al.(2022)Lipman, Chen, Ben-Hamu, Nickel, and Le]{lipman2022flow}
Yaron Lipman, Ricky~TQ Chen, Heli Ben-Hamu, Maximilian Nickel, and Matt Le.
\newblock Flow matching for generative modeling.
\newblock \emph{arXiv preprint arXiv:2210.02747}, 2022.

\bibitem[Liu et~al.(2025{\natexlab{a}})Liu, Liu, Liang, Yuan, Liu, Zheng, Wu, Wang, Qin, Xia, et~al.]{liu2025improving}
Jie Liu, Gongye Liu, Jiajun Liang, Ziyang Yuan, Xiaokun Liu, Mingwu Zheng, Xiele Wu, Qiulin Wang, Wenyu Qin, Menghan Xia, et~al.
\newblock Improving video generation with human feedback.
\newblock \emph{arXiv preprint arXiv:2501.13918}, 2025{\natexlab{a}}.

\bibitem[Liu et~al.(2025{\natexlab{b}})Liu, Ma, Li, Chen, Liu, He, and Wu]{liu2025phantom}
Lijie Liu, Tianxiang Ma, Bingchuan Li, Zhuowei Chen, Jiawei Liu, Qian He, and Xinglong Wu.
\newblock Phantom: Subject-consistent video generation via cross-modal alignment.
\newblock \emph{arXiv preprint arXiv:2502.11079}, 2025{\natexlab{b}}.

\bibitem[Liu et~al.()Liu, Gong, et~al.]{liuflow}
Xingchao Liu, Chengyue Gong, et~al.
\newblock Flow straight and fast: Learning to generate and transfer data with rectified flow.
\newblock In \emph{The Eleventh International Conference on Learning Representations}.

\bibitem[Mou et~al.(2025)Mou, Wu, Wu, Guo, Zhang, Cheng, Luo, Ding, Zhang, Li, et~al.]{mou2025dreamo}
Chong Mou, Yanze Wu, Wenxu Wu, Zinan Guo, Pengze Zhang, Yufeng Cheng, Yiming Luo, Fei Ding, Shiwen Zhang, Xinghui Li, et~al.
\newblock Dreamo: A unified framework for image customization.
\newblock \emph{arXiv preprint arXiv:2504.16915}, 2025.

\bibitem[Oquab et~al.(2023)Oquab, Darcet, Moutakanni, Vo, Szafraniec, Khalidov, Fernandez, Haziza, Massa, El-Nouby, et~al.]{oquab2023dinov2}
Maxime Oquab, Timoth{\'e}e Darcet, Th{\'e}o Moutakanni, Huy Vo, Marc Szafraniec, Vasil Khalidov, Pierre Fernandez, Daniel Haziza, Francisco Massa, Alaaeldin El-Nouby, et~al.
\newblock Dinov2: Learning robust visual features without supervision.
\newblock \emph{arXiv preprint arXiv:2304.07193}, 2023.

\bibitem[Oquab et~al.(2024)Oquab, Darcet, Moutakanni, Vo, Szafraniec, Khalidov, Fernandez, Haziza, Massa, El-Nouby, et~al.]{oquab2024dinov2}
Maxime Oquab, Timoth{\'e}e Darcet, Th{\'e}o Moutakanni, Huy Vo, Marc Szafraniec, Vasil Khalidov, Pierre Fernandez, Daniel Haziza, Francisco Massa, Alaaeldin El-Nouby, et~al.
\newblock Dinov2: Learning robust visual features without supervision.
\newblock \emph{Transactions on Machine Learning Research Journal}, 2024.

\bibitem[Peng et~al.(2024)Peng, Cui, Tang, Qi, Dong, Bai, Han, Ge, Zhang, and Xia]{peng2024dreambench++}
Yuang Peng, Yuxin Cui, Haomiao Tang, Zekun Qi, Runpei Dong, Jing Bai, Chunrui Han, Zheng Ge, Xiangyu Zhang, and Shu-Tao Xia.
\newblock Dreambench++: A human-aligned benchmark for personalized image generation.
\newblock \emph{arXiv preprint arXiv:2406.16855}, 2024.

\bibitem[Polyak et~al.(2024)Polyak, Zohar, Brown, Tjandra, Sinha, Lee, Vyas, Shi, Ma, Chuang, et~al.]{polyak2024movie}
Adam Polyak, Amit Zohar, Andrew Brown, Andros Tjandra, Animesh Sinha, Ann Lee, Apoorv Vyas, Bowen Shi, Chih-Yao Ma, Ching-Yao Chuang, et~al.
\newblock Movie gen: A cast of media foundation models.
\newblock \emph{arXiv preprint arXiv:2410.13720}, 2024.

\bibitem[Radford et~al.(2021)Radford, Kim, Hallacy, Ramesh, Goh, Agarwal, Sastry, Askell, Mishkin, Clark, et~al.]{radford2021learning}
Alec Radford, Jong~Wook Kim, Chris Hallacy, Aditya Ramesh, Gabriel Goh, Sandhini Agarwal, Girish Sastry, Amanda Askell, Pamela Mishkin, Jack Clark, et~al.
\newblock Learning transferable visual models from natural language supervision.
\newblock In \emph{International conference on machine learning}, pages 8748--8763. PmLR, 2021.

\bibitem[Sand-AI(2025)]{magi1}
Sand-AI.
\newblock Magi-1: Autoregressive video generation at scale, 2025.
\newblock URL \url{https://static.magi.world/static/files/MAGI_1.pdf}.

\bibitem[Schuhmann et~al.(2022)Schuhmann, Beaumont, Vencu, Gordon, Wightman, Cherti, Coombes, Katta, Mullis, Wortsman, et~al.]{schuhmann2022laion}
Christoph Schuhmann, Romain Beaumont, Richard Vencu, Cade Gordon, Ross Wightman, Mehdi Cherti, Theo Coombes, Aarush Katta, Clayton Mullis, Mitchell Wortsman, et~al.
\newblock Laion-5b: An open large-scale dataset for training next generation image-text models.
\newblock \emph{Advances in neural information processing systems}, 35:\penalty0 25278--25294, 2022.

\bibitem[Seawead et~al.(2025)Seawead, Yang, Lin, Zhao, Lin, Ma, Guo, Chen, Qi, Wang, et~al.]{seawead2025seaweed}
Team Seawead, Ceyuan Yang, Zhijie Lin, Yang Zhao, Shanchuan Lin, Zhibei Ma, Haoyuan Guo, Hao Chen, Lu~Qi, Sen Wang, et~al.
\newblock Seaweed-7b: Cost-effective training of video generation foundation model.
\newblock \emph{arXiv preprint arXiv:2504.08685}, 2025.

\bibitem[Shao and Cui(2023)]{link1}
Shihao Shao and Qinghua Cui.
\newblock 1st solution in google universal image embedding.
\newblock \url{https://www.kaggle.com/datasets/louieshao/guieweights0732}, april 2023.

\bibitem[Wang et~al.(2025{\natexlab{a}})Wang, Ai, Wen, Mao, Xie, Chen, Yu, Zhao, Yang, Zeng, Wang, Zhang, Zhou, Wang, Chen, Zhu, Zhao, Yan, Huang, Feng, Zhang, Li, Wu, Chu, Feng, Zhang, Sun, Fang, Wang, Gui, Weng, Shen, Lin, Wang, Wang, Zhou, Wang, Shen, Yu, Shi, Huang, Xu, Kou, Lv, Li, Liu, Wang, Zhang, Huang, Li, Wu, Liu, Pan, Zheng, Hong, Shi, Feng, Jiang, Han, Wu, and Liu]{wan2025}
Ang Wang, Baole Ai, Bin Wen, Chaojie Mao, Chen-Wei Xie, Di~Chen, Feiwu Yu, Haiming Zhao, Jianxiao Yang, Jianyuan Zeng, Jiayu Wang, Jingfeng Zhang, Jingren Zhou, Jinkai Wang, Jixuan Chen, Kai Zhu, Kang Zhao, Keyu Yan, Lianghua Huang, Mengyang Feng, Ningyi Zhang, Pandeng Li, Pingyu Wu, Ruihang Chu, Ruili Feng, Shiwei Zhang, Siyang Sun, Tao Fang, Tianxing Wang, Tianyi Gui, Tingyu Weng, Tong Shen, Wei Lin, Wei Wang, Wei Wang, Wenmeng Zhou, Wente Wang, Wenting Shen, Wenyuan Yu, Xianzhong Shi, Xiaoming Huang, Xin Xu, Yan Kou, Yangyu Lv, Yifei Li, Yijing Liu, Yiming Wang, Yingya Zhang, Yitong Huang, Yong Li, You Wu, Yu~Liu, Yulin Pan, Yun Zheng, Yuntao Hong, Yupeng Shi, Yutong Feng, Zeyinzi Jiang, Zhen Han, Zhi-Fan Wu, and Ziyu Liu.
\newblock Wan: Open and advanced large-scale video generative models.
\newblock \emph{arXiv preprint arXiv:2503.20314}, 2025{\natexlab{a}}.

\bibitem[Wang et~al.(2025{\natexlab{b}})Wang, Ai, Wen, Mao, Xie, Chen, Yu, Zhao, Yang, Zeng, et~al.]{wang2025wan}
Ang Wang, Baole Ai, Bin Wen, Chaojie Mao, Chen-Wei Xie, Di~Chen, Feiwu Yu, Haiming Zhao, Jianxiao Yang, Jianyuan Zeng, et~al.
\newblock Wan: Open and advanced large-scale video generative models.
\newblock \emph{arXiv preprint arXiv:2503.20314}, 2025{\natexlab{b}}.

\bibitem[Wang et~al.(2024{\natexlab{a}})Wang, Shi, Ou, Chen, Lin, Wang, Jiang, Yang, Zheng, Tao, et~al.]{wang2024koala}
Qiuheng Wang, Yukai Shi, Jiarong Ou, Rui Chen, Ke~Lin, Jiahao Wang, Boyuan Jiang, Haotian Yang, Mingwu Zheng, Xin Tao, et~al.
\newblock Koala-36m: A large-scale video dataset improving consistency between fine-grained conditions and video content.
\newblock \emph{arXiv preprint arXiv:2410.08260}, 2024{\natexlab{a}}.

\bibitem[Wang et~al.(2024{\natexlab{b}})Wang, Liu, Lin, Yan, Chen, Low, Hoang, Wu, Liew, Yan, et~al.]{wang2024magicvideo}
Weimin Wang, Jiawei Liu, Zhijie Lin, Jiangqiao Yan, Shuo Chen, Chetwin Low, Tuyen Hoang, Jie Wu, Jun~Hao Liew, Hanshu Yan, et~al.
\newblock Magicvideo-v2: Multi-stage high-aesthetic video generation.
\newblock \emph{arXiv preprint arXiv:2401.04468}, 2024{\natexlab{b}}.

\bibitem[Winter et~al.(2024)Winter, Shul, Cohen, Berman, Pritch, Rav-Acha, and Hoshen]{winter2024objectmate}
Daniel Winter, Asaf Shul, Matan Cohen, Dana Berman, Yael Pritch, Alex Rav-Acha, and Yedid Hoshen.
\newblock Objectmate: A recurrence prior for object insertion and subject-driven generation.
\newblock \emph{arXiv preprint arXiv:2412.08645}, 2024.

\bibitem[Wu et~al.(2025)Wu, Zhu, and Shou]{wu2025automated}
Weijia Wu, Zeyu Zhu, and Mike~Zheng Shou.
\newblock Automated movie generation via multi-agent cot planning.
\newblock \emph{arXiv preprint arXiv:2503.07314}, 2025.

\bibitem[Xu et~al.()Xu, Xie, Tan, Huang, Howes, Sharma, Li, Ghosh, Zettlemoyer, and Feichtenhofer]{xudemystifying}
Hu~Xu, Saining Xie, Xiaoqing Tan, Po-Yao Huang, Russell Howes, Vasu Sharma, Shang-Wen Li, Gargi Ghosh, Luke Zettlemoyer, and Christoph Feichtenhofer.
\newblock Demystifying clip data.
\newblock In \emph{The Twelfth International Conference on Learning Representations}.

\bibitem[Xu et~al.(2024)Xu, Zhang, Liew, Yan, Liu, Zhang, Feng, and Shou]{xu2024magicanimate}
Zhongcong Xu, Jianfeng Zhang, Jun~Hao Liew, Hanshu Yan, Jia-Wei Liu, Chenxu Zhang, Jiashi Feng, and Mike~Zheng Shou.
\newblock Magicanimate: Temporally consistent human image animation using diffusion model.
\newblock In \emph{Proceedings of the IEEE/CVF Conference on Computer Vision and Pattern Recognition}, pages 1481--1490, 2024.

\bibitem[Yang et~al.(2024{\natexlab{a}})Yang, Yang, Zhang, Hui, Zheng, Yu, Li, Liu, Huang, Wei, et~al.]{yang2024qwen2}
An~Yang, Baosong Yang, Beichen Zhang, Binyuan Hui, Bo~Zheng, Bowen Yu, Chengyuan Li, Dayiheng Liu, Fei Huang, Haoran Wei, et~al.
\newblock Qwen2. 5 technical report.
\newblock \emph{arXiv preprint arXiv:2412.15115}, 2024{\natexlab{a}}.

\bibitem[Yang et~al.(2024{\natexlab{b}})Yang, Teng, Zheng, Ding, Huang, Xu, Yang, Hong, Zhang, Feng, et~al.]{yang2024cogvideox}
Zhuoyi Yang, Jiayan Teng, Wendi Zheng, Ming Ding, Shiyu Huang, Jiazheng Xu, Yuanming Yang, Wenyi Hong, Xiaohan Zhang, Guanyu Feng, et~al.
\newblock Cogvideox: Text-to-video diffusion models with an expert transformer.
\newblock \emph{arXiv preprint arXiv:2408.06072}, 2024{\natexlab{b}}.

\bibitem[Zeng et~al.(2024)Zeng, Wei, Zheng, Zou, Wei, Zhang, and Li]{zeng2024make}
Yan Zeng, Guoqiang Wei, Jiani Zheng, Jiaxin Zou, Yang Wei, Yuchen Zhang, and Hang Li.
\newblock Make pixels dance: High-dynamic video generation.
\newblock In \emph{Proceedings of the IEEE/CVF Conference on Computer Vision and Pattern Recognition}, pages 8850--8860, 2024.

\bibitem[Zhong et~al.(2025)Zhong, Yang, Teng, Gu, and Li]{zhong2025concat}
Yong Zhong, Zhuoyi Yang, Jiayan Teng, Xiaotao Gu, and Chongxuan Li.
\newblock Concat-id: Towards universal identity-preserving video synthesis.
\newblock \emph{arXiv preprint arXiv:2503.14151}, 2025.

\end{thebibliography}

\clearpage
\beginappendix
\begin{figure}[t]
  \centering
  \includegraphics[width=\linewidth]{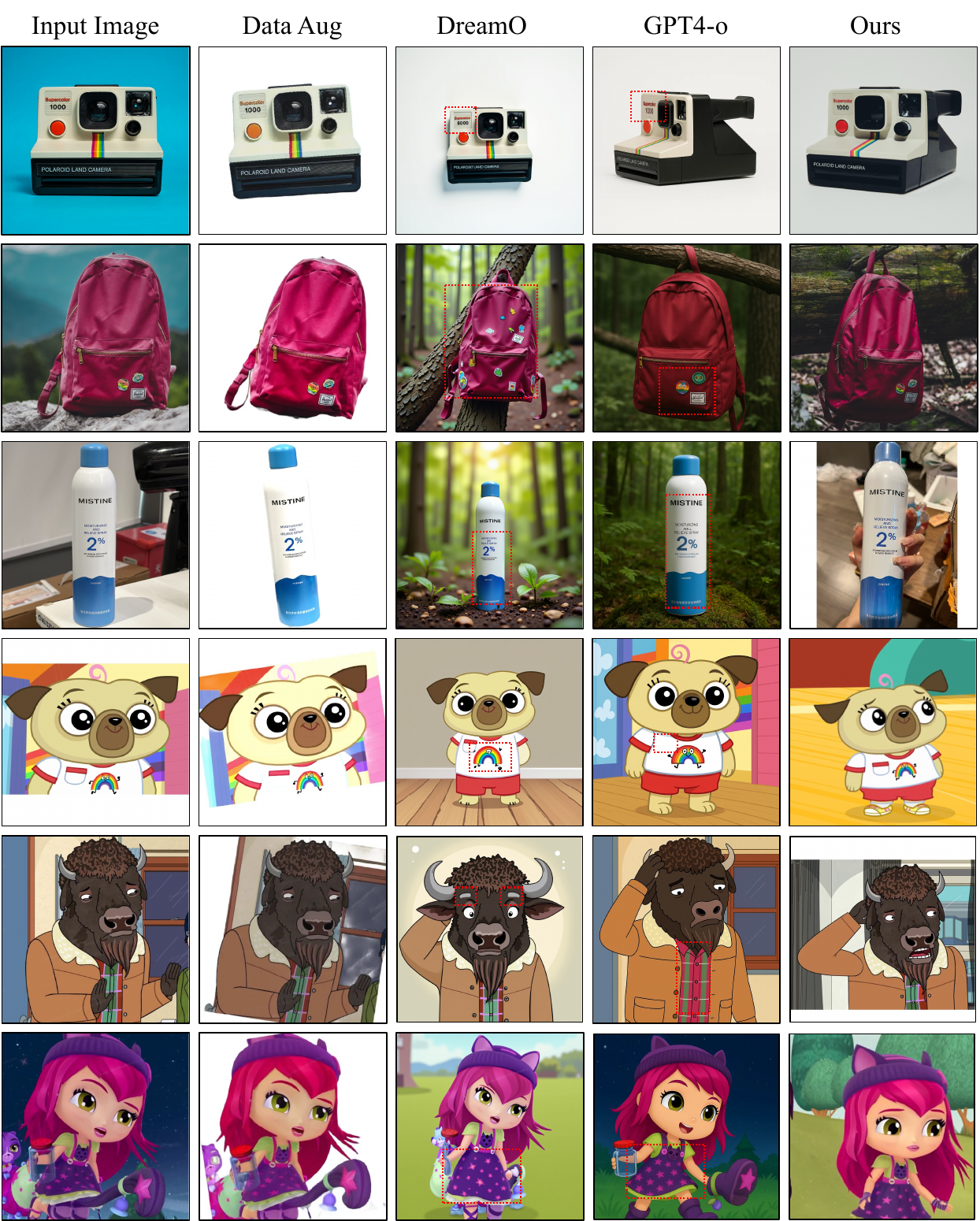}
  \caption{Comparison of reference-data construction strategies.
Naïve data augmentation (Data Aug) offers only limited variation, while directly adopting state-of-the-art IP-consistent generators (e.g., DreamO, GPT-4o) can introduce appearance inconsistencies, highlighted in the figure. Best viewed in color with zoom.}
  \label{fig:diff_data}
\end{figure}
\section{The Limitations of Synthetic Data}\label{sec:synthetic_limitation}
We try two SOTA models, GPT4o and DreamO \cite{mou2025dreamo} to generate consistent subjects on different context. The results in Fig.\ref{fig:diff_data} shows these models could still generated inconsistent subject. However our real cross-pair data construction pipeline could provide exactly same subjects on different context. 

\section{User study}
To evaluate the generated videos under different training data regimes, we conducted a user study comparing four settings: in-pair training, in-pair with data augmentation, face-level cross-pair training, and our proposed full-object cross-pair approach. We ask six participants, each of whom independently evaluated 50 video groups, containing four videos generated in the different training settings. For each group, participants were asked to select the best video in terms of overall visual quality, subject consistency, and alignment with textual prompts. As summarized in Table~\ref{tab:user_study}, our method was overwhelmingly preferred, receiving \textbf{76\%} of the votes. In contrast, all other baselines received less than 12\%, highlighting the effectiveness of our cross-pair training design in producing videos that are faithful to the intent of users.
\begin{table}[tbh]
\centering
\renewcommand{\arraystretch}{1.2}
\begin{tabular}{cccc}
\hline
 In-pair  & In-pair + Data Aug  & Face Cross-pair & Ours  \\
\hline
 6\% & 11\% & 7\% & 76\%  \\
\hline
\end{tabular}
\caption{User study on the best video selection based on overall visual quality, subject consistency, and text alignment across different training data settings. }
\label{tab:user_study}
\end{table}

\section{Broader Impact}
Our work aims to enhance identity consistency and contextual diversity in subject-to-video generation, thereby improving the controllability and realism of AI-generated content. This technology holds promise for a wide range of applications, including personalized media creation, digital asset generation, and educational or entertainment content production.

Nonetheless, we recognize the potential social risks associated with realistic identity-preserving video synthesis. Such capabilities may be misused for malicious purposes, including the creation of deepfakes, impersonation, or the dissemination of misinformation. We therefore emphasize the importance of responsible research and deployment practices. In particular, we encourage the use of watermarking, provenance tracking, and informed consent mechanisms to ensure ethical and transparent use—especially in scenarios involving human likeness or identity-sensitive content.
\end{document}